\documentclass[letterpaper, 10 pt, journal, twoside]{IEEEtran}

  \usepackage{pgfplots}
  \pgfplotsset{compat=newest}
  \usetikzlibrary{plotmarks}
  \usetikzlibrary{arrows.meta}
  \usepgfplotslibrary{patchplots}
  \usepackage{grffile}
  \usepackage{amsmath}
  \usepackage{blindtext}
  \usepackage[bottom]{footmisc}
  \usepackage{mathrsfs}
  \usepackage{lipsum}
 \usepackage{url}
 \usepackage{array}
\usepackage{booktabs}
\usepackage{float}
\usepackage{multirow}
\usepackage{cite}

\newlength\fwidth
 \usepackage[linesnumbered,ruled]{algorithm2e}

\usepackage{graphicx}
\usepackage{framed}
\usepackage{subcaption}
\usepackage{indentfirst}
\usepackage[]{algorithm2e}
\usepackage{multirow}
\usepackage{makecell}

\usepackage{amsmath, amssymb,amsthm}
\usepackage{color}
\usepackage{booktabs}
\usepackage{tabularx}

\usepackage{comment}
\usepackage{vector}
\usepackage{hyperref}
\usepackage{cleveref}

\crefname{figure}{Fig.}{Fig.}
\Crefname{figure}{Fig.}{Fig.}
\usepackage[export]{adjustbox}

\begin{document}

\title{Towards Efficient Human-Robot Collaboration with\\ Robust Plan Recognition and Trajectory Prediction}

\author{Yujiao Cheng$^{1}$, Liting Sun$^{1}$, Changliu Liu$^{2}$, and Masayoshi Tomizuka$^{1}$% <-this % stops a space
%\thanks{This work was supported by National Science Foundation (Award \#1734109).}
\thanks{Manuscript received: September, 10, 2019; Revised December, 8, 2019; Accepted February, 2, 2020.}%Use only for final RAL version
\thanks{This paper was recommended for publication by Editor Youngjin Choi upon evaluation of the Associate Editor
	and Reviewers’ comments.
%This work was supported by (organizations/grants which supported the work.)
} %Use only for final RAL version
\thanks{$^{1}$First Author, Second Author and Fourth Author are with School of Engineering, Mechanical Engineering Department, University of California, Berkeley
        {\tt\footnotesize yujiaocheng, liting, tomizuka@berkeley.edu}}%
\thanks{$^{2} $Third Author is with Robotics Institute, Carnegie Mellon University, Pittsburgh
        {\tt\footnotesize cliu6@andrew.cmu.edu}}%
\thanks{Digital Object Identifier (DOI): see top of this page.}
}

\markboth{IEEE Robotics and Automation Letters. Preprint Version. Accepted February, 2020}
{Cheng \MakeLowercase{\textit{et al.}}: Towards Efficient Human-Robot Collaboration} 

\maketitle

\begin{abstract}
Human-robot collaboration (HRC) is becoming increasingly important as the paradigm of manufacturing is shifting from mass production to mass customization. The introduction of HRC can significantly improve the flexibility and intelligence of automation. To efficiently finish tasks in HRC systems, the robots need to not only predict the future movements of human, but also more high-level plans, i.e., the sequence of actions to finish the tasks. However, due to the stochastic and time-varying nature of human collaborators, it is quite challenging for the robot to efficiently and accurately identify such task plans and respond in a safe manner. To address this challenge, we propose an integrated human-robot collaboration framework. Both plan recognition and trajectory prediction modules are included for the generation of safe and efficient robotic motions. Such a framework enables the robots to perceive, predict and adapt their actions to the human's work plan and intelligently avoid collisions with the human. Moreover, by explicitly leveraging the hierarchical relationship between plans and trajectories, more robust plan recognition performance can be achieved. Physical experiments were conducted on an industrial robot to verify the proposed framework. The results show that the proposed framework could accurately recognize the human workers' plans and thus significantly improve the time efficiency of the HRC team even in the presence of motion classification noises. 
\end{abstract}


\section{Introduction}

% \begin{figure}[b]
% \begin{center}
% \includegraphics[width=.7\linewidth]{FutureFactory_LaptopAssembly.pdf}
% \caption{Flexible production lines in the future, involving both human-robot co-operation and co-inhabitance.}
% \label{fig: factory}
% \end{center}
% \end{figure}

\IEEEPARstart{A}{s} the emphasis of manufacturing is shifting from mass production to mass customization, the demands for flexible automation keep increasing. Human-robot collaboration (HRC), as an effective and efficient way to enhance the flexibility, has attracted lots of attention both in industry and academia in the past decade. 
% As shown in Fig.~\ref{fig: factory}, 
% As shown in Fig.~\ref{fig: factory}, 
The idea of HRC is to let robots work safely and collaboratively with humans in a shared space. To achieve this, robots should be equipped with various capabilities from fundamental skills, such as perception of human activities, to higher-level social skills, including reasoning about intentions and collaboration~\cite{thomaz2016computational}.

Collaboration between humans and intelligent robots can be categorized into three levels: 1) low-level collision avoidance, 2) middle-level efficient cooperation with task plan recognition and trajectory prediction, and 3) high-level collaboration mode selection and automatic task assignments. Many researches have been conducted for all these three levels. The first category regards human as moving obstacles and designs algorithms to let the robot avoid collisions with human. The third category, e.g., \cite{devin2017decisions}, studies the task assignment algorithms in peer-to-peer human-robot interaction where humans and robots work as partners.

In this paper, we focus on the second category %: middle-level efficient cooperation with human trajectory prediction and plan recognition. 
which includes three key elements. The first one is human trajectory prediction which aims to predict the continuous human movement used for safe robot trajectory planning. It is different from discrete intention recognition such as \cite{perez2015fast, pellegrinelli2016human, liu2016goal, koppula2016anticipatory}. Many approaches have been proposed for continuous trajectory prediction, from early attempts such as Kalman filter and particle filter~\cite{kohler1997using}~\cite{bruce2004better} to recent effort such as recurrent neural networks \cite{ghosh2017learning}, inverse reinforcement learning \cite{sun_probabilistic_2018}~\cite{Sun2019ITSC} and semi-adaptable neural network \cite{cheng2019human}. 
A human-aware robotic system which incorporated both motion predictions and trajectory planning has also been presented in \cite{unhelkar2018human}.

The second element is human plan recognition. It aims to recognize what plan the human is doing given the observed trajectories and his/her influence to the objects. %It requires good perception of the human activity and a robot planner to adapt to the human.
Human activity/plan recognition has attracted a great amount of effort. Some work focused on deep learning frameworks with RGBD images as inputs \cite{sung2011human}~\cite{mukherjee2018human}~\cite{zhang2012rgb}. Typically, the features selected mainly focus on human, for instance, the body pose, hand positions, motion information and histogram of oriented gradients (HOG). No information about the objects of interaction are included. However, the objects can provide rich information for inferring what the human is doing via the intrinsic hierarchy among actions, motions and the objects. Hence, in this paper, we explore such hierarchy to design more robust plan recognition algorithm. 
%Note that our work is different from \cite{thomas1996hierarchical}~\cite{hayes2016autonomously} which try to construct an efficient plan representation considering the hierarchy among plans, while our work tries to explore such hierarchy in the plan recognition algorithm.

With good plan recognition module and trajectory prediction module, the third key element for efficient human-robot collaboration is the behavior/action generation for robots, i.e., the planner. Although there are many robot behavior generation approaches \cite{koppula2016anticipatory}~\cite{fiore2016planning, nikolaidis2016formalizing, nikolaidis2013human, thomaz2016computational, sun2018courteous}, most of them focus on the action level rather than the plan level. %Some works  formalized human-robot mutual adaptation. Some other works discussed in the survey paper \cite{thomaz2016computational} chose between competing human plans and executed actions from a fixed set. Our planner falls into the last category.

%Although there are some existing works on trajectory prediction or plan recognition, there is no well-established HRC framework which integrates both aspects to allow the robot to adapt to humans at both the trajectory level and task planning level. 

We propose an integrated HRC framework which includes both trajectory prediction and plan recognition. At trajectory planning level, robots take future human trajectory into account to avoid potential collisions, which improves safety. At task planning level, robots perceive the human's actions, infer the human plan and adapt to the human's actions in advance to boost the collaboration efficiency. 
\begin{comment}
we aim to improve the efficiency of human-robot collaboration while assuring safety via reliable plan recognition and trajectory prediction. To achieve this, an integrated HRC framework is proposed, which enables the robot to perceive the human's actions, infer the human's plans and adapt to the human's actions to achieve efficient and safe collaboration. 

\end{comment}
The proposed framework advantages HRC in three aspects. First, the robot is more responsive to the human's plans, particularly when there might be change of plans in the human's operations. By using our proposed plan recognition method, the robot can quickly recognize the human's new plan, and adapt its actions accordingly. Experimental results showed that the average task completion time is significantly reduced, i.e., more efficient HRC can be achieved. Second, our system is robust with respect to noises in the model inputs and errors in the intermediate steps such as motion classification. We combine a long short-term memory (LSTM) network with algorithms based on Bayesian inference instead of end-to-end learning. Moreover, a set of hierarchical relationships among trajectories, motions, actions, plans and tasks are explicitly defined and utilized in the plan recognition algorithm. This not only helps improve the robustness of the algorithm, but also reduce the dimension of the problem, and enhance its generalization ability using less data.
\begin{comment}
we might need to say how much data we have used to train this network. In this part, you might need to borrow some of your results in the semi-adaptable network.
\end{comment}
% All these advantages of the proposed framework have been verified via experiments on an industrial assembly settings in Section \ref{sec: exp}.

The key contributions of this work are:
\begin{enumerate}
    \item We propose a robot system interleaving prediction/recognition and adaptation at both trajectory and task planning level.
    \item We propose a robust plan recognition algorithm by leveraging the hierarchical relationship between the plan and the trajectories.
    \item We provide physical experiments to evaluate the whole system in a desktop assembly task using an industrial robot arm. Results demonstrate that the system improves the efficiency of HRC, and the plan recognition algorithm is robust in the presence of noises such as false motion classification.
\end{enumerate}

%The remainder of this paper is organized as follows. In section \ref{sec: framework}, an integrated HRC framework is proposed. In section \ref{sec: pr}, we present the detailed plan recognition algorithm, followed by a description of a trajectory prediction method in section \ref{sec: prediction}. Section \ref{sec: planner} describes the mechanism of the planner. Experiments are shown in section \ref{sec: exp} and section \ref{sec: conclusion} concludes the paper.

\section{An integrated Plan Recognition and Trajectory Prediction Framework}
\label{sec: framework}
In this work, we focus on enabling better HRC systems via plan recognition and trajectory prediction. The terminologies we used in this paper are defined as follows:
\begin{itemize}
	\item \textbf{Trajectory}: a time series of the joint positions of an agent (either a human or a robot) in Cartesian space. It represents the continuous movements of an agent.
	\item \textbf{Motion}: A discrete variable/label to represent different types/patterns of trajectories. For instance, typical motions in factory scenarios include ``Fetching'', ``Picking'', ``Screwing'' and ``Taping''. Different trajectories can be generated to perform the same motion pattern.
	\item \textbf{Action}: A paired discrete variable/label including a motion pattern and the target object to act on, i.e., $\text{action}{=}\{\text{motion}, \text{object}\}$. For example, we can define ``$\text{action 1}{=}\{\text{Fetching}, \text{a screwdriver}\}$'', ``$\text{action 2}{=}\{\text{Taping}, \text{a bunch of cables}\}$'' and so on.
	\item \textbf{Subtask}: A subtask is an element of completing a larger \emph{task} (defined below), whose initial states and the goal states do not depend on other subtasks. It might be implemented with several sequences of actions, depending on their orders.

	\item \textbf{Plan}: A plan is comprised of a sequence of ordered subtasks. It represents the preferences to finish a \emph{task} (defined below). Different orders of actions in different plans come from either orders of subtasks, or the orders of actions within subtasks.

	\item \textbf{Task}: A task represents the work to be conducted by agents. It specifies the initial states, the goal states and the participants. A task can be decomposed into a set of subtasks and executed via a variety of plans.
\end{itemize}
\begin{figure}[ht]
	\begin{center}
		\includegraphics[width=\linewidth]{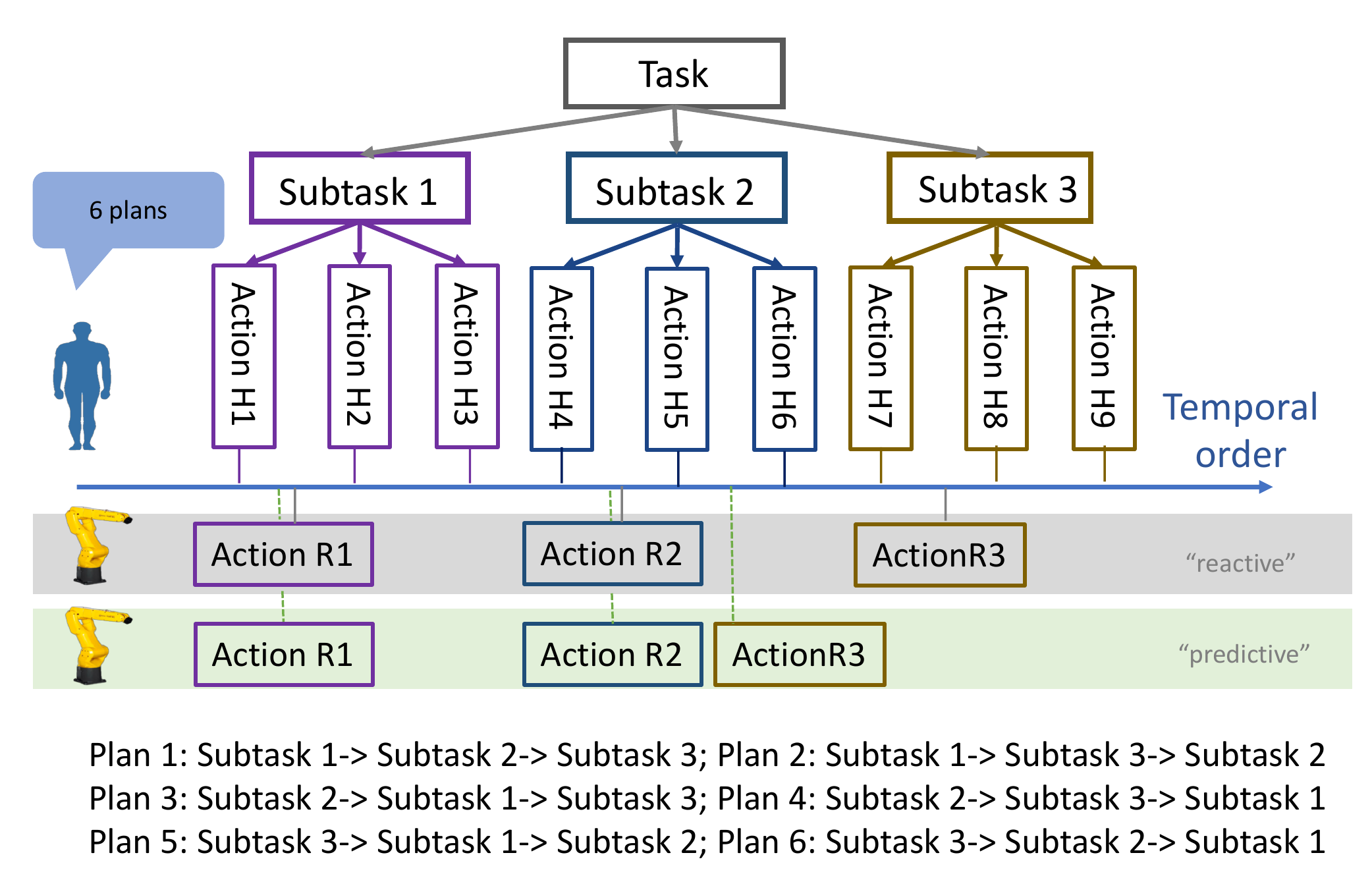}
		\caption{A hierarchical and temporal decomposition of a task.}
		\label{fig: task}
	\end{center}
\end{figure}

We use Fig.~\ref{fig: task} and a \textbf{desktop assembly} example to illustrate the hierarchical relationship of the terminologies. 
A task (``desktop assembly") can be decomposed into three subtasks (``installing a CPU fan", ``installing a system fan", ``taping cables"). Suppose each of them has a unique action order, then the permutation of three subtasks totally generates six different plans, all of which are stored in the plan library as action\footnote{In the desktop assembly example, Action H1-H9 are ``fetching the cpu fan", ``receiving the screwdriver A", ``screwing the cpu fan", ``fetching the system fan", ``receiving the screwdriver B", ``screwing the system fan", ``taping the cables", ``receiving scissors", and ``cutting the tape". Action R1-R3 are ``delivering screwdriver A", ``delivering screwdriver B" and ``delivering scissors"}  sequences of the human and the robot. Furthermore, within each action, the motion can be executed by infinite many (theoretically) trajectories.
Trajectory prediction is to forecast the future movement of the human, thus the robot can make safe trajectory planning avoiding potential collisions.
Plan recognition is to choose the correct plan in human's mind, which is to choose the predefined action sequence in the plan library. 
As shown in Fig.~\ref{fig: task}, without plan recognition, the robot (the ``reactive" robot) can only acquire its next action after the human finishes some key actions (such as Action H1, Action H4 and Action H7), while with a plan recognition, the robot (the ``predictive" robot) can foresee the future actions of the human and execute its following actions in advance to boost the efficiency of the collaboration. 

However, human plans are not directly observable for the robot. The only observable variables of humans are their trajectories, which means that the robot has to infer and reason about the probable plans by observing the trajectories of humans. Such diversity, un-observability, and time-varying characteristic of the human plans create great challenges for the HRC systems. It requires the robot to 1) quickly and reliably recognize the plans of humans, and 2) responsively adapt its own behavior in a safe and predictive manner to ensure efficient and seamless collaboration.

\begin{figure}[h]
\begin{center}
\includegraphics[width=\linewidth]{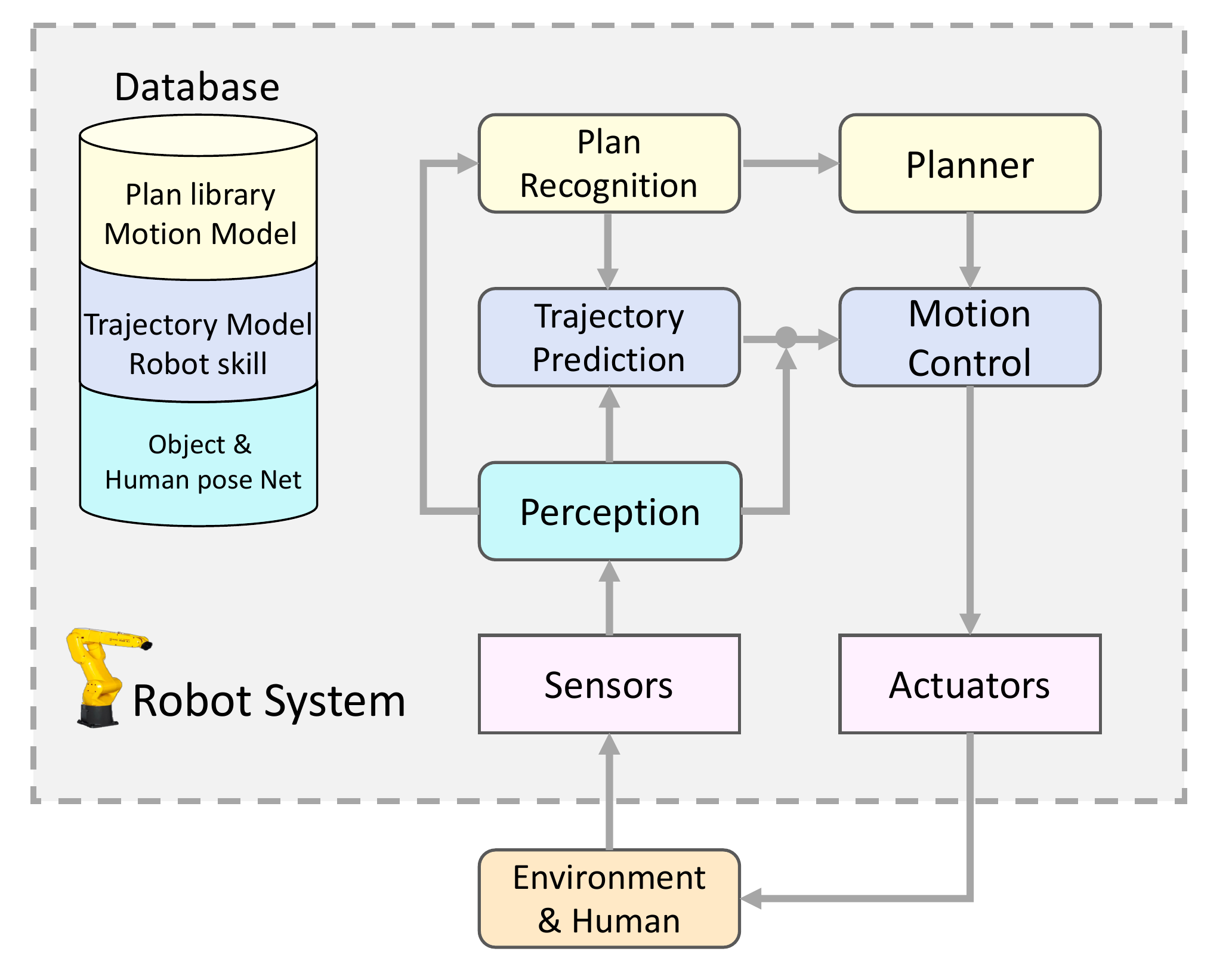}
\caption{The architecture of the proposed integrated HRC framework}
\label{fig: archi}
\end{center}
\end{figure}
To address these two challenges, we propose an integrated HRC framework, the architecture of which is shown in Fig.~\ref{fig: archi}. It includes both offline database and online modules. Online modules include a perception module (sensors and perception algorithms), a plan recognition module, a trajectory prediction module, a planner, a motion control module and the actuators (the robot). 
%The goal of each online module is given below.
\textit{Perception Module} takes visual information as inputs and outputs the 3D positions of objects as well as the 3D human poses. 
\textit{Plan Recognition Module} is a key module in our proposed framework. It aims to identify the action being executed by the human and infer human's plan by observing the trajectories of their key joints. 
The action estimate will be sent to both the planner module and the trajectory prediction module.
\textit{Planner Module} assigns the next action (a motion-object pair) to the robot based on the current states, the recognized plan and the current action of the human. The action command from the planner will be sent to the motion control module.
\textit{Trajectory Prediction Module} aims to predict the future trajectories of the human. Instead of directly predicting the future trajectories based on only current and historical human trajectories, we leverage the action labels from the plan recognition module.
\textit{Motion Control Module} includes two controllers: an efficiency controller and a safety controller, as in \cite{liu2018robot}. The efficiency controller is a long-term global controller to assure the efficiency of robot, and the safety controller is a short-term local controller for real-time safety under uncertainties.

\section{The Plan Recognition Algorithm}
\label{sec: pr}
%Assisting a human worker with assembly task without explicit commands requires the knowledge of human's plan.However, possible plans for a task can be a lot, and human's choice of plan is a state of mind that is immeasurable. Even human cannot make good estimation of the plan of others. Thus, we need a probabilistic method to recognize the plan from external clues. The key of external clues is actions performed by the human.We make such assumption that the human's actions are purely driven by the plan in mind. 
%Human is also assumed to be rational, and only one plan of one task is undertaken until the task is complete. \yujiao{can delete} 
%Hence, to recognize the human's plan, we need to keep track of human's action, and then based on human action trajectory, the plan of human can be recognized.
As discussed in Section \ref{sec: framework}, to enable efficient and seamless human-robot collaboration, the robot needs to quickly and reliably recognize the plan executed by a human, and safely adapt its behavior. However, the diversity of human plans for the same task, the time-inconsistent or time-varying characteristic of humans, as well as the un-observability of plans have posed great challenges for accurate and timely plan recognition. To address these challenges, we propose a plan recognition algorithm based on both deep learning techniques and Bayesian inference. Moreover, we explicitly take advantage of the hierarchical relationships among ``trajectory", ``action", ``subtask" and ``plan", and design the plan recognition into three mutually compensated steps for better plan recognition. The four steps of plan recognition are: motion classification, target object estimation, plan inference and posterior action correction. 

The proposed plan recognition algorithm has advantages in two aspects. First, the dimension of action space is reduced via the hierarchical combination of motion and object. Notice that an action is defined as a pair of motion and object, and the set of candidate objects for different motions can be quite different. Hence, a hierarchical combination of motion classification and target object estimation can help significantly reduce the dimension of the classification problem compared to direct action classification. Second, more robust recognition performance can be achieved via the posterior update step of the action based on the plan information. With this step, prior domain knowledge in regard to the relationships of plan and actions is fully utilized to help reduce the sensitivity of learning based methods to noises.

\begin{figure}[]
\begin{center}
\includegraphics[width=\linewidth]{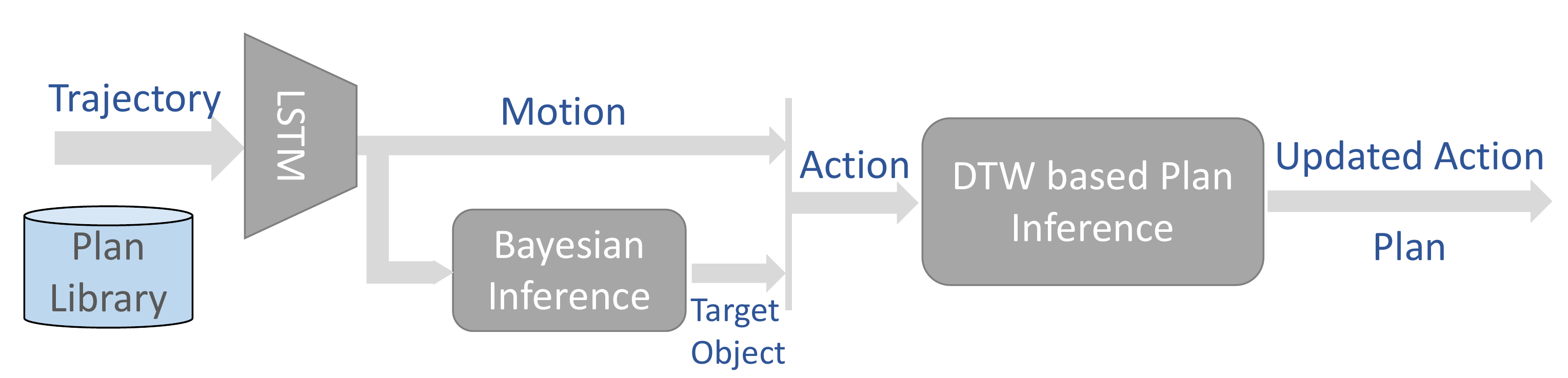}
\caption{The architecture of the plan recognition module.}
\label{fig: LSTM}
\end{center}
\end{figure}

\subsection{Motion Classification}
Motion classification aims to categorize different motions given segments of trajectories of the human's key joints. Long-short-term-memory (LSTM) neural networks have been extensively proved to be an effective approach to model the dynamics and dependencies in sequential data. %\cite{liu2017global}. 
Hence, we design a LSTM recurrent neural network for motion classification. 
% The structure of the LSTM network is depicted in Fig.~\ref{fig: LSTM}.
The input data is the human pose from the \emph{Perception} module.  
More specifically, in an assembly task, the input vector at time step $k$ is $\textbf{x}_k{=}\{\textbf{w}_k,\textbf{h}_k \}$, where $\textbf{w}_k$ is the wrist position in the world frame and $\textbf{h}_k$ are the velocities of selected key points on the human fingers.
%\yujiao{should explain?} We do not use the 3D hand keypoints feature because 2D information captures features of the motion well. 
%In this sense, we only observe one hand of the human to detect the motion, but it is easy to extend the features to whole arm and even whole body if necessary.
The output at time step $k$ is a motion label $m_k{\in}\{1,2,...n_m\}$, where $n_m{\in}\mathbb{N}$ is the number of motions.
%with a categorical probability distribution $\textbf{p}_k\in\mathbb{R}^{n_m}$ from softmax output, where $n_m\in\mathbb{N}$ is the number of motions. 
The LSTM is trained using the "Motion Model" database in Fig.~\ref{fig: archi}.
% \begin{figure}[h]
% \begin{center}
% \includegraphics[width=\linewidth]{LSTM.pdf}
% \caption{The architecture of the motion classification network}
% \label{fig: LSTM}
% \end{center}
% \end{figure}
\subsection{Target Object Estimation}
%To ensure the action detection is stable and reliable, before we continue to decide on the target object of interaction, we expect that the motion classification result satisfies two conditions:
%\begin{itemize}
%  \item The best motion has much higher probability than the second best.
%  \item The best motion lasts for a few ($>2$) timesteps.
%\end{itemize}
%First condition is to make sure that the best motion is prominent among others. If the second best is almost equally likely, that means our motion detection is not confident enough. The second condition is to enhance our confidence of the motion detection. Action lasts for a period of time, so it is reasonable to expect the motion detection to give consistent result. Once these conditions are satisfied, we believe the human motion is $m_k = r_k$ with confidence and the set of candidate objects can be easily retrieved based on affordance matrix. Let the motion indicator vector be $I_m\in R^{n_m}$, then the set of the candidate objects can be indexed by the indices of non-zeros elements of $\mathcal{A}I_m$.  If these conditions are not satisfied, we will set $m_k = m_{k-1}$, since we believe it is most likely that the human does not change actions. 

Given the classified motion labels and a history of human pose, 
%the set of object candidates $\mathcal{O}$ is updated, and then 
Bayesian inference is commonly used to update the beliefs on different target objects, e.g. \cite{das2018cross}.

Let $o_k$ be an object at time step $k$, $\mathcal{O}$ be the object set, $m_{1:k}$ be the historical motion labels, and $h_{1:k}$ be the historical human poses. Then we need to obtain the robot's beliefs on the object, i.e., a probability $P(o_k|h_{1:k},m_{1:k})$.
Applying the Markov assumption, the following equation holds:

\begin{align*}
&P(o_k|h_{1:k},m_{1:k}) \propto \\
%P(h_k, m_k | o_k, h_{1:k-1}, m_{1:k-1})\cdot \\
&P(m_k|o_k, h_{k-1}, m_{k-1})\cdot P(h_k|o_k, h_{k-1}, m_{k}) \\ 
&\sum_{o_{k-1}\in\mathcal{O}} P(o_k|h_{k-1}, m_{k-1}, o_{k-1}) \cdot P(o_{k-1}|h_{1:k-1},m_{1:k-1})
\end{align*}

We compute the $P(h_k|o_k, h_{k-1}, m_{k})$ with an assumption that humans are optimizing some value function as \cite{baker2009action} suggests. Then a Boltzmann policy can be applied:
$$ 
P(h_k|o_k, h_{k-1}, m_{k}) \propto \text{exp}(\beta V_g(h_{k};o_k))
$$
where $V_g$ is the value function. We model $V_g$ for each motion as a function of distance and velocity. 

To compute $P(m_k|o_k, h_{k-1}, m_{k-1})$ and $P(o_k|h_{k-1}, m_{k-1}, o_{k-1})$, we impose conditional independence assumption of $m_k$ and $h_{k-1}$ given $o_{k}$ and $m_{k-1}$, and conditional independence assumption of $o_k$ and $h_{k-1}$ given $o_{k-1}$ and $m_{k-1}$. Then, with predefined or learned models of $P(m_k|m_{k-1},o_{k})$ and $P(o_k|m_{k-1},o_{k-1})$, $P(o_k|h_{1:k},m_{1:k})$ can be updated iteratively.

\subsection{Plan Inference}
\label{sec: plan}
With results from motion classification and object estimation, we can uniquely determine a sequence of actions by observing the human trajectories. Note that a plan is a sequence of subtasks, and each subtask is represented by one action or an ordered sequence of actions. Hence, a plan can be uniquely represented by a temporal sequence of actions. Therefore, we first build a plan library offline in the \emph{Database} where each plan is represented by a reference sequence of actions. Then we utilize the reference sequences to online infer potential plans based on Bayes' rule,
\begin{align*}
P(g|a_{1:k}) \propto P(a_{1:k}|g)P(g),
\end{align*}
where $P(g)$ is a prior belief of plan $g$, and $P(g|a_{1:k})$ is a posterior belief based on the likelihood of observed action sequence $a_{1:k}$ given plan $g$. Similarly, with Boltzmann policy, the likelihood of the action trajectory can be defined as
$$
P(a_{1:k}|g) \propto \text{exp}(-d(a_{1:k}; g)),
$$
where the function $d$ is a distance function measuring the similarity between observed action sequence (A, namely $a_{1:k}$) and the reference action sequence (R) of the plan $g$. The larger the distance is, the less likely the human is following the plan \cite{kapur1992entropy}. 
We adopt the open-end dynamic time warping (OE-DTW) algorithm \cite{tormene2009matching} to calculate $d$. 
This algorithm is to best match the query sequence to a reference sequence and calculate the dissimilarity between the matched portion. 
% The idea of the algorithm is to compute lengths of both the matchedthe prefix of reference best matched by the input and the dissimilarity between the matched portion. This is different from dynamic time warping (DTW), which is to align two complete sequences, so the start end points must aligned between two sequences. 
%In our case, the reference comes from stretching the reference action sequence by the minimum time for each action. The input is the action trajectory. The size of the prefix of the reference best matched indicates the posterior action.
\begin{comment}
\begin{figure}[]
\begin{center}
\includegraphics[height = 0.7\linewidth, width=.95\linewidth]{dtw.pdf}
\caption{The OE-DTW distance.}
\label{fig: dtw}
\end{center}
\end{figure}
\end{comment}
% A brief review of the algorithm is as follows. 
% Suppose that a human plan in \emph{Database} is represented by a temporal action sequence \{Action H1, Action H2, ... , Action Hi, ...  Action Hp\}, and we enforce that Action Hi should take at least $t_i$ time steps. Then the reference action sequence (R) is of length $n_r=\sum_{i=1}^pt_i$ by duplicating Action Hi for $t_i$ times.
Given a reference time series $R = (r_1, r_2,  ..., r_{N})$ and a query sequence $A = (a_1, a_2, ..., a_M)$,
the OE-DTW distance between A and R is calculated via minimizing the dynamic time warping distances (DTW) between A and any references $R^j$ truncated from reference R at point $j = 1:N$. 
$$
D_{OE}(A,R) = \min_{j =1,...,N} D_{DTW}(A, R^{j}).
$$

Here is a short introduction to DTW.
The indices of the two series will be mapped through $\phi_t$ and $\psi_t$, $t = 1,2,...,T$, that satisfy the following constraints \cite{tormene2009matching}:
\begin{itemize}
  \item Boundary condition: $\phi_1=1$, $ \psi_1 = 1$ and $\phi_T=N$, $ \psi_T= M$
  \item Monotonic conditions: $\phi_{t-1} \leq \phi_t$ and $\psi_{t-1} \leq \psi_t$ 
  \item Continuity conditions: $\phi_t - \phi_{t-1} \leq 1$, and $\psi_t - \psi_{t-1} \leq 1$  % the distance will push j ahead
  \item Local slope constraints: certain step patterns are allowed.
\end{itemize}
%$\hat{\Phi}$

The optimal $\hat{\Phi} = (\hat{\phi}_t,\hat{ \psi}_t)$ minimizes the distance between the two warped time series:
$$
(\hat{\phi}_t, \hat{\psi}_t) = arg\min_{\phi_t, \psi_t} \sum_{t=1}^T \frac{d(r_{\phi_t}, a_{\psi_t})m_{t, \Phi}}{\sum_tm_{t,\Phi}},
$$
where $d(\cdot, \cdot)$ is any distance function and $m_{t,\Phi}$ is a local weighting coefficient.  Therefore, the dynamic time warping distance between A and R is 
$$
D_{DTW}(A,R) = \sum_{t=1}^T \frac{d(r_{\hat{\phi}_t}, a_{\hat{\psi}_t})m_{t, \Phi}}{\sum_tm_{t,\Phi}}. 
$$

\subsection{Posterior Action Correction}
As we obtain the posterior estimate of the plan $g^*$, the best matched reference sequence $R^*{}^{j^*}$ is also obtained. 
We correct the action label estimate $a_k$ by retrieving the action in the best matched reference plan as follows,
$$
a_k^{post} = R^*(j^*)=r^*_{j^*}.
$$
This step is of key importance to reduce the sensitivity of the learning models to noises, so that the robustness of the plan recognition can be improved. The effectiveness of this step is verified in experiments.
  
\section{Human Trajectory Prediction}\label{sec: prediction}
To avoid collisions between a human and a robot, the future human trajectory is required to be considered in the safety controller to generate safety constraints. 
We leverage two inputs: 1) the human pose estimates from \emph{perception module} and 2) the action labels from the \emph{plan recognition module}. The human transition model is approximated by a feedforward neural network, the output layer of which is adapted online using recursive least square parameter adaptation algorithm (RLS-PAA) to address the challenges regarding the time-varying characteristic of human trajectories.
For more details of this approach, one can refer to our previous work \cite{cheng2019human}.

\begin{comment}
\begin{equation}
	\mathbf{x}(k+1) =\Phi_k \theta_k+w_k.\label{eq:Transformed LTV}
\end{equation}

\begin{align}
	\hat{\mathbf{x}}\left(k+1|k\right) =&\Phi_k\hat{\theta}_k,\label{eq:prediction}\\
	\tilde{\mathbf{x}}\left(k+1|k\right) =&\Phi_k\tilde{\theta}_k+w_k.
\end{align}

\begin{align}
	& X_{\tilde{\theta}\tilde{\theta}}\left(k+1\right) \nonumber\\
	= & F_k\Phi^{T}_kX_{\tilde{x}\tilde{x}}\left(k+1|k\right)\Phi_kF_k\nonumber  -X_{\tilde{\theta}\tilde{\theta}}(k)\Phi^{T}_k\Phi_kF_k \\
	&-F_k\Phi^{T}_k\Phi_kX_{\tilde{\theta}\tilde{\theta}}(k)\nonumber 
	 +E\left[\tilde{\theta}_{k+1}\right]\Delta\theta^{T}_k\\
	 &+\Delta\theta_kE\left[\tilde{\theta}_{k+1}\right]^{T} -\Delta\theta_k\Delta\theta_k^{T}+X_{\tilde{\theta}\tilde{\theta}}(k).\label{eq:parameter-estimation-error-covariance}
\end{align}
\end{comment}

\begin{algorithm}[h!]
\SetAlgoLined
\textbf{Input} plan library $\mathcal{Q}$; motion models $\mathcal{M}$; trajectory prediction models $f^*$; a set of objects $\mathcal{O}$\\
 %\hspace*{\algorithmicindent} \textbf{Output} 
\SetKwInOut{}{}
\textbf{Init}: RobotIsDoing = \{\}, NewHumanPose = \{\}, RobotActionBuffer=\{\}\;
 \While{true}{
  NewHumanPose = getValidPoseFromPerception()\;
  \If{notEmpty(NewHumanPose)}{
   record historical human joint trajectory $h_{1:k}$\;
   $m_k$$\leftarrow$ MotionClassification($h_{1:k}$, $\mathcal{M}$)\; 
   $o_k$ $\leftarrow$ TargetObjectEstimation ($m_{1:k}$, $o_{1:k-1}$, $h_{1:k}$)\;
   $h_{k+1:k+M}$ $\leftarrow$ TrajectoryPrediction($h_{1:k}$, $f^*$, $m_k$, $o_k$)\;
   obtain $a_k=\{m_k, o_k\}$ \; 
   generate action trajectory $a_{1:k}$\; 
   $p(\textbf{g}|a_{1:k})$, $a_k^{post}$ $\leftarrow$ OEDTWPlanInference($\mathcal{Q}$, $a_{1:k}$)\;
   $\hat{g}^{[1]}$, \ $\hat{g}^{[2]}$ $\leftarrow$
   the best and second best plan estimates\;
%   $\hat{g}^{(2)}$\leftarrow argmax\ $p(\textbf{g}^{-(1)}\vert a_{1:k})$ \ls{what does this mean? second largest}\;\\
  \eIf{$p(\hat{g}^{[1]}|a_{1:k})>$Threshold}{
  RobotActionBuffer$\leftarrow$ nextActionSequence($\hat{g}^{[1]}$,$a_k^{post}$,$\mathcal{Q}$) \; 
  }
  { 
  action1 $\leftarrow$ nextAction($\hat{g}^{[1]}$,$a_k^{post}$,$\mathcal{Q}$)\;
  action2 $\leftarrow$ nextAction($\hat{g}^{[2]}$,$a_k^{post}$, $\mathcal{Q}$)\; 
  \If{action1==action2}{RobotActionBuffer$\leftarrow${action1}\;
  }
  }
  \If{ notEmpty(RobotActionBuffer)}
  {
  \If{notEmpty(RobotIsDoing)\& RobotActionBuffer\{1\}!=RobotIsDoing}{
  recover what robot is doing\;
  RobotIsDoing $\leftarrow$ RobotActionBuffer\{1\}\;
  }
  }
  }
 
  Execute(RobotIsDoing, $h_{k+1:k+M}$)\;
  \If{ActionExecutionFinished}{
  \eIf{notEmpty(nextInBuffer)}{
  RobotIsDoing $\leftarrow$ nextInBuffer\;
  }{
  RobotIsDoing = \{\}\;
  }
  }
 }
 
 \caption{Proposed HRC system} \label{algo: 1}
\end{algorithm}
\section{The Planner}\label{sec: planner}
In this section, we will present the \emph{Planner} module and the workflow of the proposed framework. 
As shown in Fig.~\ref{fig: archi}, the output of the \emph{Plan recognition} module is sent to the \emph{Planner} to generate commands for the \emph{Motion control} module. With the identified plan and the human action estimate, the \emph{Planner} module acquires the next action of the robot from the plan library, and sets the goal states of the \emph{Motion control} module to generate safe and executable trajectories. 
%For example, if the robot's next action is to bring the screwdriver to the human. The \emph{Planner} module will set a sequence of the goal states in the \emph{Motion control} module as ''end effector to be at the position where the screwdriver is located", "grasp the screwdriver", ''end effector to a position near the human's hands". Meanwhile, the predicted trajectory of human joints will also be sent to \emph{Motion control} module to generate safety constraints.

Since the plan recognition results are probabilistic, we need to design a decision-making mechanism to decide on the robot actions. There are two cases that we might encounter. The first one is that the probability of one plan is prominently higher than that of others. 
 This gives the \emph{Planner} a clear idea about the plan the human is executing, and it can directly acquire all the following actions for the robot from the plan library. 
 The other case is where there are two or more candidate plans with similar probabilities from the plan recognition algorithms. Under this situation, the \emph{Planner} will look at the two most likely plans, and find out whether the next action of the robot for each plan is the same. If the next action is the same, \emph{Planner} will directly let the robot execute it. Otherwise, the \emph{Planner} will wait and collect the human's information to clear out the confusion.

The \emph{Planner} also takes changes of plan into consideration. If the robot's next action is not consistent with what the robot is doing, it will recover the current action and responsively adjust its action. For example, if the robot is delievering screwdriver to the human, while suddenly the next action becomes bringing the scissors. The robot will put back the screwdriver if it already grabs it, and go to scissors immediately. The pseudo-code for the workflow of our proposed system is presented as Algorithm \ref{algo: 1}.

\section{Experiments}\label{sec: exp}
\subsection{A Desktop Assembly Scenario} \label{sec: sce}
We evaluate our proposed framework in a desktop assembly task in industrial settings.
The task of the HRC team is to assemble a desktop (desktop assembly example explained in Section \ref{sec: framework}). This task can be decomposed into three un-ordered subtasks: installing a CPU fan, installing a system fan and taping cables. Each subtask is implemented by only one action sequence. Thus, by a hierarchical decomposition as shown in Fig.~\ref{fig: task}, there are at most six different plans to finish the task. The robot is designed to assist the human by delivering necessary tools to the human as he/she needs. Definitions of actions for the human and the robot can be found in Section \ref{sec: framework}.

As depicted in the Fig.~\ref{fig: task}, a predictive robot with an effective plan recognition will recognize the human's plan in the second subtask, and proactively execute the following actions in sequence. For example, if the human is doing the first plan, namely, installing a CPU fan, installing a system fan, and finally taping the cables. We expect that correct plan is inferred when the human is fetching the system fan and then the robot will execute the following actions ("delivering the screwdriver" and "delivering the scissors") in sequence.

% \begin{figure}[t]
% \begin{center}
% \includegraphics[width=.7\linewidth]{setup.pdf}
% \caption{The Experiment Setup in the Benchmark Scenario.}
% \label{fig: setup}
% \end{center}
% \end{figure}

\subsection{Experiment Design}
\subsubsection{Hypothesis} We evaluate the effectiveness of the proposed plan recognition and trajectory prediction framework by verifying the following three hypotheses. 
\begin{itemize}
    \item H1: The proposed framework is a safe HRC framework.
    \item H2: The proposed framework improves the efficiency of the HRC team.
    \item H3: The performance of the proposed framework is robust to noises or errors caused by some intermediate steps such as motion classification.
    {\item H4: The human subjects are more satisfied with our collaborative robots than with a responsive robot in terms of some criteria.}
\end{itemize}
\subsubsection{Experiment setup} We test our system on an industrial robot FANUC LR Mate 200iD/7L. A Kinect V2 for windows is placed close to the table on which a robot arm and a human worker do task together. Some necessary tools lie in the tool area and a CPU fan and a system fan are in the part area. We conducted experiments with human in the loop. Eight human subjects participated in the experiments.

\subsubsection{Manipulated variables} To evaluate the effectiveness of the proposed framework, we manipulated two controlled variables in our experiments: \textit{plan recognition schemes} and \textit{trajectory prediction schemes}. We define ``plan recognition = 0" as no plan recognition, ``plan recognition = 1" as recognition ground truths provided by human subjects, and ``plan recognition = 2" as the recognition results generated by the proposed algorithm in Section \ref{sec: pr}. When ``plan recognition = 0", the robot is completely reactive, meaning that it receives the information of the human action after the human completes the action. The robot only starts to move once it detects the human's actions, and it can only collaborate with human within subtasks. When ``plan recognition = 1", the robot has perfect plan knowledge, and it moves based on the ground truths of human's actions and plan. When ``plan recognition = 2", the proposed algorithm will let the robot automatically identify the human's actions and infer about the potential plan, so that the human and robot can collaborate across subtasks. In addition, we define ``trajectory prediction = 0" as no predictions of human trajectory, and ``trajectory prediction = 1" as prediction via our proposed method in Section \ref{sec: prediction}. By manipulating the two variables, we have six groups of experiments. Under each group, every human subject performs the task using any plan for three times. Thus, there will be $24$ trials in each experiment group and in total we collect $144$ trials for all groups.

\subsubsection{Dependent measures} 
To quantify the safety of the proposed framework, we measure the minimum distance between a human subject and the robot during the entire task in each trial. Smaller the minimum distance, the less safe it is for the human.
For efficiency, we use a timer to keep track of the task completion time. The timer starts when a human subject starts to move, and ends when the task is finished. 
To quantify the plan recognition performance, we calculate the plan recognition accuracy and the action recognition accuracy which is intermediate result of the plan recognition module. Plan and action recognition accuracy is the percentage of plan estimates and action estimates that conform to the true values labeled by human subjects. 
Notice that plan recognition takes place at every time step, and there might be multiple plan labels in the early phase. As long as the estimate is one of the labels, it is regarded as correct. 
{As for the measurement of human's satisfaction with our collaborative robots, we ask the eight human subjects to rate the following statements on Likert scale from $1$ (strongly disagree) to $5$ (strongly agree), similar to \cite{koppula2016anticipatory}: 1. The robot was collaborative and helped; 2. The robot did the right thing at the tight time; 3. I am satisfied working with the robot; 4. I will work with this robot again in the future. }

\subsection{Implementation Details}
The motion model was approximated by a one-layer LSTM network. We offline collected $200$ trials of ''fetching", ''receiving", ''screwing", and ''taping", $50$ trials for each class. The hyper-parameter was chosen by cross validation, and we picked $60$ to be the number of hidden units. 

Human transition model in the module of human trajectory was approximated by a fully connected neural network. We used the data set as described above. We predicted the future trajectory of $1 s$ given past trajectory of $1 s$. The number of layers was set to $3$ and the number of hidden units was set to $40$ also by cross validation.

\subsection{Results}
\textbf{{H1}}: Through extensive experiments, the minimum distances between the human subjects and the robot were, respectively, $34.9{\pm}3 cm$ and $36.0{\pm}2 cm$ with ``trajectory prediction = 0" (72 trials) and ``trajectory prediction = 1" (72 trials). Under both experiment conditions, the minimum distances were within safe distance and no collisions happened. The minimum distances in experiments with ``trajectory prediction = 1" were larger than those in the experiments with ``trajectory prediction = 0", although not significantly differ\footnote{We use paired t-test for all the statistical tests.} ($p{>}0.05$). There are two possible reasons: 1) in our scenario, the robot and the human worked in a relatively large space, so predictions of the human trajectory did not make much difference to the robot trajectory planning; 2) human subjects were conservative around the robot, and they intentionally kept a safe distance from the robot. 

To show that trajectory prediction module in our framework improves safety, we did additional tests where the human subjects aggressively move towards the robot end effector (We have safety mechanism which immediately stops the robot if contact happens.). $100$ trials with trajectory prediction and $100$ trials without trajectory prediction were collected. The collision rates were 0/100 and 64/100, respectively. Such results qualitatively showed that the trajectory prediction module can improve safety.

\begin{comment}
\begin{figure}[]
\begin{center}
\includegraphics[width=\linewidth]{avoidance.png}
\caption{A robot avoids collisions with a human based on trajectory prediction}
\label{fig: avoidance}
\end{center}
\end{figure}
\yujiao{time constraint least time }
\end{comment}

\textbf{{H2}}: The task completion time for different plan recognition schemes were recorded among trials with the eight human subjects. Without plan recognition (``plan recognition = 0"), the average task completion time is $90.0 \pm 10.9 s$, which is the longest. With our proposed plan recognition algorithm (``plan recognition = 2"), the average task completion time is $64.6\pm 10.6 s$, which is reduced by $29.1\%$. Thus, the proposed framework with plan recognition significantly improves ($p<0.01$) the efficiency of the HRC team compared to the system without the plan recognition.
As a matter of fact, our system can achieve similar performance as a system with perfect plan recognition (``plan recognition = 1") with $1.2 s$ more average task completion time and larger variance.

\textbf{{H3}}: This hypothesis can be proved via quantitative results in Table \ref{tab:Tab1}. 
One can see that although some motion classification accuracy is low, %\footnote{The online motion classification deteriorated mainly because there were a lot of random or transition movements during the experiments that do not belong to any class of the four motions. },
the plan recognition accuracy still remains high.
The Pearson product-moment correlation coefficient for the two variables is $-0.11 (p<0.01)$, which indicates weak correlation. This means that the overall performance of the proposed plan recognition algorithm is not sensitive to the errors in the intermediate LSTM step. This is mainly benefiting from the Bayesian inference step and the dynamic time warping step. These two steps serve as a low pass filter, eliminating the wrong motion estimates. Besides, the plan is actually estimated by the nearest neighbor in DTW step, and the six plans as action sequences lie sparsely in an increasingly high dimensional space, and they get farther away from each other over time. As long as the estimates do not deviate from the true point too much, the plan estimate should be correct. 
\begin{table}[h!]
   \caption{Quantitative experimental results for recognition} 
   \label{tab:Tab1}
   \small % text size of table content
   \centering % center the table
   \begin{tabular}{lccr} % alignment of each column data
   \toprule[\heavyrulewidth]\toprule[\heavyrulewidth]
   \textbf{Subjects} & \textbf{MC accuracy(\%)} & \textbf{PR accuracy (\%)} \\ 
   \midrule
   Subject 1 (6 trials)& 85.3 $\pm$ 1.3   &   97.6 $\pm$ 0.3 \\
   Subject 2 (6 trials) &  69.6 $\pm$ 15 & 98.5 $\pm$ 0.5 \\
    Subject 3 (6 trials) &81.4 $\pm$ 1.1 & 97.4 $\pm$ 1.2\\
    Subject 4 (6 trials)   &87.9 $\pm$ 7.0 & 97.6 $\pm$ 1.1\\
    Subject 5 (6 trials) &  79.4 $\pm$ 6.5  & 96.4 $\pm$ 1.8\\
    Subject 6 (6 trials) & 80.7 $\pm$ 12.0 & 97.9 $\pm$ 0.1\\
    Subject 7 (6 trials) & 84.8 $\pm$ 9.5 & 97.6 $\pm$ 0.6\\
    Subject 8 (6 trials) & 85.8 $\pm$ 8.8 & 90.5 $\pm$ 0.3\\
   \bottomrule[\heavyrulewidth] 
   \end{tabular}
\end{table}

\begin{figure}[h!]
\begin{center}
\includegraphics[width=.7\linewidth]{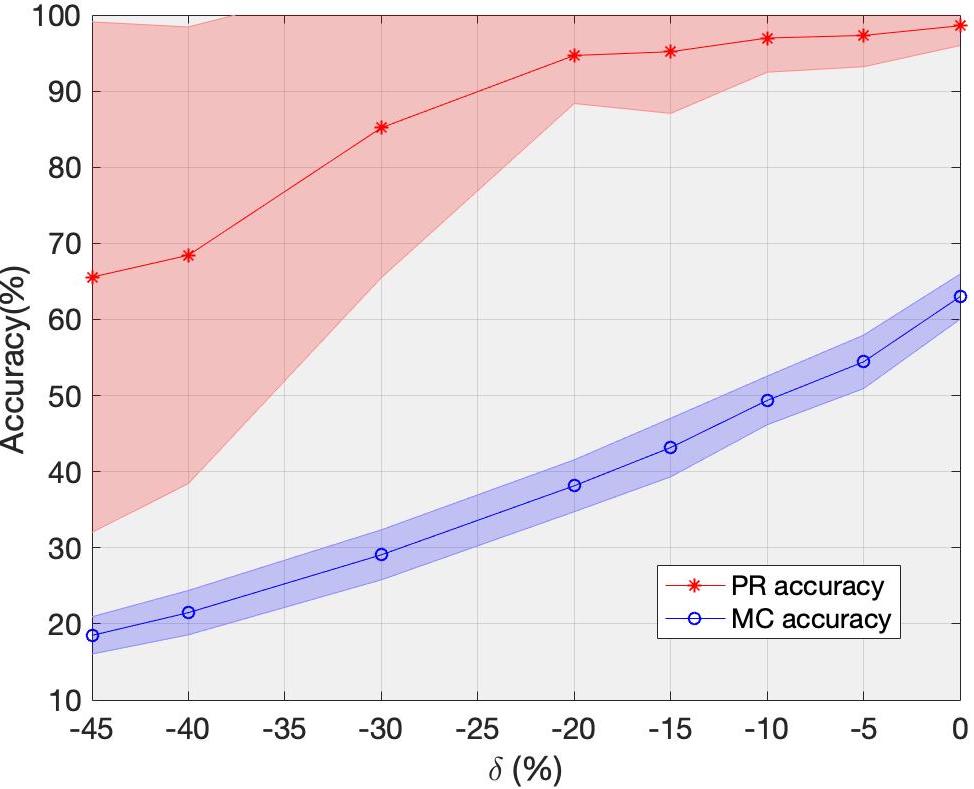}
\caption{Simulations for plan recognition accuracy when reducing the motion classification accuracy.}
\label{fig: simulation}
\end{center}
\end{figure}

To further validate the robustness brought by the DTW step, we assumed that the target object detection was perfect and did simulations by varying the motion classification (MC) performance and then tested the plan recognition (PR) accuracy. 
First, we obtained the true positive rates for each motion throughout all the experiments with ``plan recognition = 2": $83.4\%$ for ``screwing", $64.2\%$ for ``fetching", $59.1\%$ for ``receiving", and $84.2\%$ for ``taping". Then we varied each true positive rate by $\delta (\%)$, which took values of $0, -5, -10, -15, -20, -30, -40$, and $-45$. Based on these sets of true positive rates, we simulated $15$ trials of action sequences for each of the six plans in the desktop assembly task, and so we had $90$ trials for each $\delta$. As we can see in the Fig.~\ref{fig: simulation}, when $\delta$ is equal to $-30$ (true positive rate for each motion is $53.4\%,24.2\%,29.1\%,54.2\%$) and the overall motion classification is $30\%$, the plan recognition accuracy remains higher than $85\%$, which shows robustness of our plan recognition to the motion accuracy.

\begin{figure}[t!]
\begin{center}
\includegraphics[width=1\linewidth]{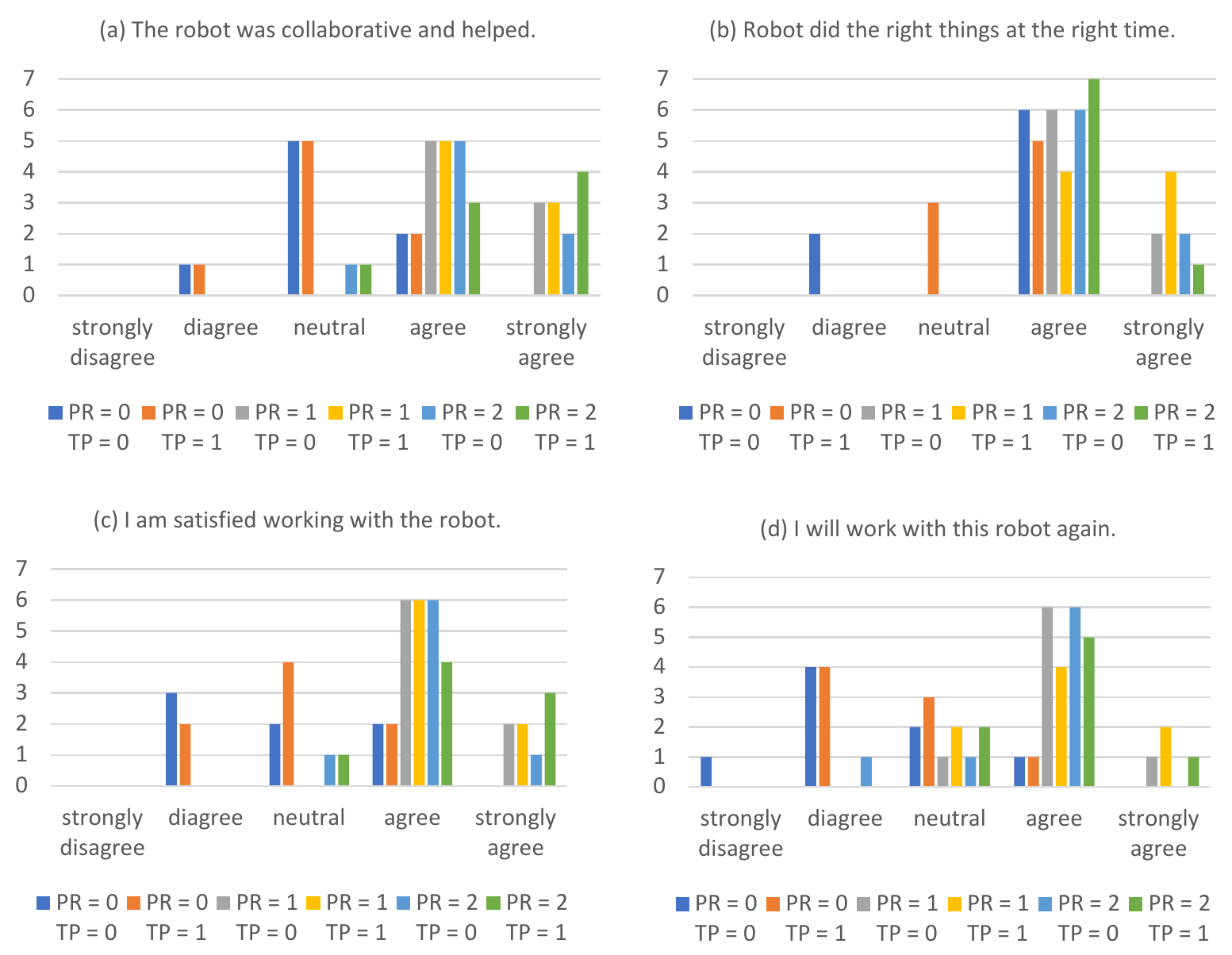}
\caption{Human subjects‘ ratings for six types of robots on four different criteria. PR is short for plan recognition and TP is short for trajectory prediction.}
\label{fig: user}
\end{center}
\end{figure}

\textbf{{H4}}: Fig.~\ref{fig: user} shows the comparison of human subjects' ratings for the six types of robots on four criteria mentioned above. Human subjects rated the robot with our proposed plan recognition algorithm (``plan recognition = 2'') significantly higher ($p<0.01$) than the robot without plan recognition (``plan recognition = 0'') on all four criteria.
Between robots with ground truths of plan recognition (``plan recognition = 1'') and robots with our plan recognition algorithm (``plan recognition = 2''), there is no significant difference ($p>0.05$) on all the criteria except for the criteria ``The robot did the right things at the right time'' ($p = 0.04$). Furthermore, there is also no significant difference ($p>0.05$) between the robots with trajectory prediction (``trajectory prediction'' = 1) and the robots without trajectory prediction (``trajectory prediction'' = 0). This might be because the trajectory prediction is too short to influence the human's feedback. Recalling the fact that "plan recognition" has significant influence, we can see that human care more about efficient plan recognition.

\textbf{Aside}: The sensitivity of threashold is not obvious. We designed a new experiment by varying the threshold value in Algorithm 1 on the experiment data we collected. It was found that the plan recognition results remained the same when the threshold was dropped from $0.70$ to $0.58$, where $0.70$ is the threshold we used in other experimental results. 

\section{Extension to Multiple Tasks}
Our plan recognition algorithm also works to distinguish different tasks. We test our algorithm on three different tasks in the CAD-60 Cornell Activity Datasets \cite{sung2011human}: cooking (stir), opening a pill container and drinking water. Table \ref{tab:Tab2} shows the comparison of the results in the ``new person'' setting using our algorithm and the two-layered maximum entropy Markov model (MEMM) method in \cite{sung2011human}. We can see that our plan recognition algorithm can also achieve very high accuracy compared to the approach in \cite{sung2011human}. This verifies our claim that exploiting the rich object information can help improve the task/plan recognition performance.

In addition, compared to end-to-end leanring, our algorithm advantages in two aspects: 1) the learning process is easier, since a hierarchical combination of motion classification and target estimation reduces the dimension of the classification problem; 2) the learning pipeline is more interpretable and predictable. As a cost, however, the proposed method requires stronger prior knowledge, i.e., all possible plans of the new task should be predefined offline, which might be hard when the task is complicated.

\begin{comment}
\begin{table}[t]
	\caption{Human subjects' ratings} 
	\label{tab:Tab3}
	\small % text size of table content
	\centering % center the table
	\begin{tabular}{lccccr} % alignment of each column data
		\toprule[\heavyrulewidth]\toprule[\heavyrulewidth]
		\textbf{PR, TP} & \textbf{criteria 1} & \textbf{criteria 2} & \textbf{criteria 3}& \textbf{criteria 4}\\ 
		\midrule
		0, 0 & 3.1 $\pm$ 0.6   &   3.5 $\pm$ 0.9 &   2.7 $\pm$ 0.9 &   2.4 $\pm$ 0.9\\
		0, 1 &  3.1 $\pm$ 0.6 & 3.6 $\pm$ 0.5 & 3 $\pm$ 0.8 & 2.6 $\pm$ 0.7\\
		1, 0 &  4.4 $\pm$ 0.5  & 4.2 $\pm$ 0.5 & 4.2 $\pm$ 0.5 &4 $\pm$ 0.5\\
		1, 1 & 4.4 $\pm$ 0.5 & 4.5 $\pm$ 0.5 & 4.4 $\pm$ 0.5 &4 $\pm$ 0.8\\
		2, 0 & 4.1 $\pm$ 0.6 & 4.3 $\pm$ 0.5& 4.1 $\pm$ 0.6 & 3.6 $\pm$ 0.7\\
		2, 1   &4.4 $\pm$ 0.7 & 4.1 $\pm$ 0.4 & 4.1 $\pm$ 0.6 & 3.9 $\pm$ 0.6\\
		\bottomrule[\heavyrulewidth] 
	\end{tabular}
\end{table}

\end{comment}

\begin{comment}
\begin{figure}[h]
\begin{center}
\includegraphics[width=.95\linewidth]{table_res.pdf}
\caption{Result table for three tasks.}
\label{fig: table}
\end{center}
\end{figure}
\end{comment}

\begin{table}[t]
   \caption{The result table for the three tasks} 
   \label{tab:Tab2}
   \footnotesize % text size of table content
   \centering % center the table
   \begin{tabular}{|c|c|c|c|c|c|c|} % alignment of each column data
   \hline
   \multirow{2}{*}{\textbf{Tasks}} & \multicolumn{2}{c|}{\textbf{Precision}} & \multicolumn{2}{c|}{\textbf{Recall}}& \multicolumn{2}{c|}{{$\textbf{F}_{0.5}$}}\\
    \cline{2-7}
    & ours & {MEMM} & ours & {MEMM} & ours & {MEMM}\\
   \hline
   \makecell{Drinking \\water} & 97.0 & 87.9 & 95.4 & 80.8 & 96.7 & 86.4\\
   \hline
   \makecell{Cooking \\(stirring)} & 88.9 & 65.5 & 98.4 & 43.9 & 90.6 & 59.7\\
   \hline
   \makecell{Opening pill \\container} & 84.3 & 86.4 & 87.0 & 58.0 & 84.4 & 78.7\\
    \hline
    \end{tabular}
\end{table}

\section{Conclusion}\label{sec: conclusion}
In this paper, we proposed an integrated framework for human-robot collaboration, including both the plan recognition and trajectory prediction. By explicitly leveraging the hierarchical relationships among plans, actions and trajectories, we designed a robust plan recognition algorithm based on neural networks and Bayesian inference. Experiments with human in the loop were conducted on an industrial assembly task. The results showed that with our proposed framework, the efficiency and safety of the human-robot collaboration can be improved. The average task completion time was reduced by $29.1\%$. Moreover, the proposed plan recognition algorithm was robust and reliable. Correct plan recognition was achieved even when the motion labels via neural networks are of low accuracy. The trajectory prediction module also enhanced the safety of the human by keeping a safe distance between the human and the robot, which is shown in our experiment video. 

We verified the effectiveness of the proposed algorithms on both an designed computer-assembly experiment and the CAD60 dataset, with comparison to the MEMM approach. More extensive comparison studies will be explored in the future to fully investigate the advantages of the proposed work, with integration with different human-robot collaboration frameworks.

\bibliographystyle{IEEEtran}
\bibliography{references}
\end{document}